\documentclass{article}

\usepackage{PRIMEarxiv}

\usepackage[utf8]{inputenc} 
\usepackage[T1]{fontenc}    
\usepackage{hyperref}       
\usepackage{url}            
\usepackage{booktabs}       
\usepackage{amsfonts}       
\usepackage{microtype}      
\usepackage{lipsum}
\usepackage{fancyhdr}       
\usepackage{graphicx}       
\usepackage{subfig}
\graphicspath{{media/}}     

\def\eg{\emph{e.g.}}

\def\etal{\emph{et al.}}

\title{SDOF-Tracker: Fast and Accurate Multiple Human Tracking by Skipped-Detection and Optical-Flow}

\author{
  Hitoshi Nishimura \\
  KDDI Research, Inc.\\
  \texttt{ht-nishimura@kddi-research.jp} \\
   \And
  Satoshi Komorita \\
  KDDI Research, Inc.\\
  \texttt{sa-komorita@kddi-research.jp} \\
   \And
  Yasutomo Kawanishi \\
  RIKEN, Nagoya University, KDDI Research, Inc.\\
  \texttt{yasutomo.kawanishi@riken.jp} \\
   \And
  Hiroshi Murase \\
  Nagoya University, KDDI Research, Inc.\\
  \texttt{murase@nagoya-u.jp} \\
}

\begin{document}
\maketitle

\begin{abstract}
Multiple human tracking is a fundamental problem for scene understanding.
Although both accuracy and speed are required in real-world applications, recent tracking methods based on deep learning have focused on accuracy and require substantial running time.
This study aims to improve running speed by performing human detection at a certain frame interval because it accounts for most of the running time.
The question is how to maintain accuracy while skipping human detection.
In this paper, we propose a method that complements the detection results with optical flow, based on the fact that someone's appearance does not change much between adjacent frames.
To maintain the tracking accuracy, we introduce robust interest point selection within human regions and a tracking termination metric calculated by the distribution of the interest points.
On the MOT20 dataset in the MOTChallenge, the proposed SDOF-Tracker achieved the best performance in terms of the total running speed while maintaining the MOTA metric.
Our code is available at \url{https://github.com/hitottiez/sdof-tracker}.
\end{abstract}

\section{Introduction}
Scene understanding from a video is one of the biggest challenges in computer vision.
Humans are often the center of attention in a scene, and tracking them in a video is a fundamental problem.
Multiple human tracking is the task of detecting the positions of multiple humans while maintaining their identities (IDs) over an image sequence.
In real-world applications such as surveillance, autonomous vehicles, and marketing, both {\it accuracy} and {\it speed} need to be sufficiently high.
In crowded scenes such as large stations, stadiums, and plazas, it often fails to detect humans, leading to ID switches.
ID switch is a serious problem because it can lead to a misunderstanding of human behaviors.
As well as the accuracy, the running speed is crucial in real-world applications. 
For example, the real-time recognition of suspicious behavior is essential in surveillance or for autonomous vehicles.

With the development of deep learning technology, the accuracy of human detection has been significantly improved, and the tracking-by-detection approach has become mainstream in recent years~\cite{Wojke2017simple,nishimura2020multiple,urbann2021online,braso2020learning,dai2021learning,papakis2021gcnnmatch}.
The approach achieves human tracking by detecting humans with a human detector and associating the detection results using a similarity metric.
The main advantage of this approach is that it is easy to determine the start and end of tracking even under occlusions and frame in/out.
Most of methods in this approach detect humans by a deep learning-based detector and extract features from each region using another deep learning model.
However, human detection and feature extraction take a lot of time; hence a rich computational resource is required for real-time tracking.
Some methods tackle this problem by simultaneous human detection and feature extraction with a single deep learning model~\cite{bergmann2019tracking,zhou2020tracking,karthik2020simple,zhang2020fair,ren2020tracking,xu2021transcenter}.
However, there is a limitation on the degree to which speed can be increased without losing accuracy.

This study aims to resolve the speed-accuracy trade-off problem by bypassing every-frame human detection, which is a computationally heavy task, that is, we perform it at a certain interval.
During the interval, human detection is skipped.
We named this process {\it Skipped-Detection}.
The question here is how to complement human detection in skipped frames.
We focus on the fact that someone's appearance is generally stable between adjacent frames.
In such a situation, basic features are useful to associate humans between adjacent frames at a pixel level.
Sparse optical flow~\cite{lucas1981iterative}, a type of optical flow, can estimate flow vectors at high speed by focusing on a small number of interest points.
In this paper, we use sparse optical flow to complement between skipped human detections.
The optical flow can also obtain detection results even in situations where the human detector misses someone.

Many tracking methods using optical flow have been proposed~\cite{everingham2009taking,schikora2011multi,fragkiadaki2012two,choi2015near,bullinger2017instance}, 
and they aim to improve the tracking accuracy by human detection and optical flow at every frame.
In contrast, we aim to maintain the tracking accuracy just with optical flow with the support of skipped detections.
The problem is that optical flow itself cannot determine the start and termination of tracking.
In this paper, we propose a novel human tracking method, which integrates Skipped-Detection and Optical-Flow, and name it {\it SDOF-Tracker}.
In SDOF-Tracker, tracking by optical flow starts triggered by human detection and terminates based on the variance of interest points.

Moreover, the proposed SDOF-Tracker has the following features to solve two problems we found in the preliminary experiments:
i) If setting interest points outside the human regions (background regions), humans are tracked inaccurately.
To set robust interest points, instance segmentation is performed and interest points are set inside a limited human region.
Since the instance segmentation is performed simultaneously with human detection, the running time does not increase much.
Moreover, these points are set around the head, which is less likely to be occluded.
ii) In crowded scenes, false negatives by the human detector frequently occur.
It is a crucial problem because false negatives continue to occur in subsequent frames.
To prevent false negatives, even if a target human is not detected, tracking by optical flow is continued for a while.

\section{Related Work}
In this section, we review related work on multiple human tracking based on detection and based on detection and optical flow.

\subsection{Tracking Based on Detection}
{\bf Tracking-by-Detection: }
A tracking-by-detection approach performs human tracking by detecting humans and associating the detection results using a similarity metric.
SORT~\cite{Bewley2016_sort} calculates overlap between detections and applies the Hungarian algorithm~\cite{kuhn1955hungarian} for data association.
SORT is widely used in real-world applications due to its speed, but it may fail in crowded scenes due to lack of appearance features for data association.
DeepSORT~\cite{Wojke2017simple} is an extended version of SORT.
It utilizes not only overlaps but also appearance features for data association.
MHT-MAF~\cite{nishimura2020multiple} utilizes human action features for data association.
LTSiam~\cite{urbann2021online} is based on a Siamese network, which has tandem inputs and the same weights in both branches.
MPNTrack~\cite{braso2020learning}, LPC\_MOT~\cite{dai2021learning}, and GNNMatch~\cite{papakis2021gcnnmatch} are based on a graph neural network, which captures the dependence of graphs via message passing.
However, these methods~\cite{Wojke2017simple,nishimura2020multiple,urbann2021online,braso2020learning,dai2021learning,papakis2021gcnnmatch} take a lot of time for human detection and feature extraction, so real-time tracking is unrealistic.

\noindent
{\bf Joint Detection and Tracking: }
While the tracking-by-detection approach has a two-stage structure of detection and data association, the latest approach jointly performs detection and data association in a single neural network.
Tracktor~\cite{bergmann2019tracking} can detect the position in the next frame based on the existing detector without additional training.
CenterTrack~\cite{zhou2020tracking} uses a point-by-point heatmap to predict motion, which allows for association even when someone's movement between frames is large.
SimpleReID~\cite{karthik2020simple} learns a re-identification model in an unsupervised manner.
FairMOT~\cite{zhang2020fair} is a simple model that consists of two homogeneous branches to predict pixel-wise objectness scores and re-ID features.
TBC~\cite{ren2020tracking} explicitly accounts for the object counts inferred from density maps and simultaneously solves detection and tracking.
TransCenter~\cite{xu2021transcenter} is a transformer-based architecture, which handles long-term complex dependencies by using an attention mechanism.
However, these methods are limited in terms the degree to which speed can be increased without losing accuracy because there is a trade-off between speed and accuracy. 

\subsection{Tracking Based on Detection and Optical Flow}
Human tracking methods based on detection and optical flow have been proposed in the past.
Everingham \etal proposed a method that utilizes the portion of inlier trajectories over the outliers between face detections in order to cluster them~\cite{everingham2009taking}.
Schikora \etal proposed a method that can deal with false positives and ID switches by using finite set statistics~\cite{schikora2011multi}.
Fragkiadaki \etal proposed a method that jointly optimizes detectlet classification and clustering of optical flow trajectories~\cite{fragkiadaki2012two}.
Choi~\cite{choi2015near} proposed the aggregated local flow descriptor that can accurately measure the affinity between a pair of detections.
Bullinger \etal proposed a method that exploits instance segmentation and predicts position and shape in the next frame by optical flows~\cite{bullinger2017instance}.
However, these methods require a lot of running time because they perform human detection in every frame and combine the detection result with optical flow.

\section{Proposed Method}
\begin{figure}[t]
 \begin{center}
 \includegraphics[width=0.90\linewidth]{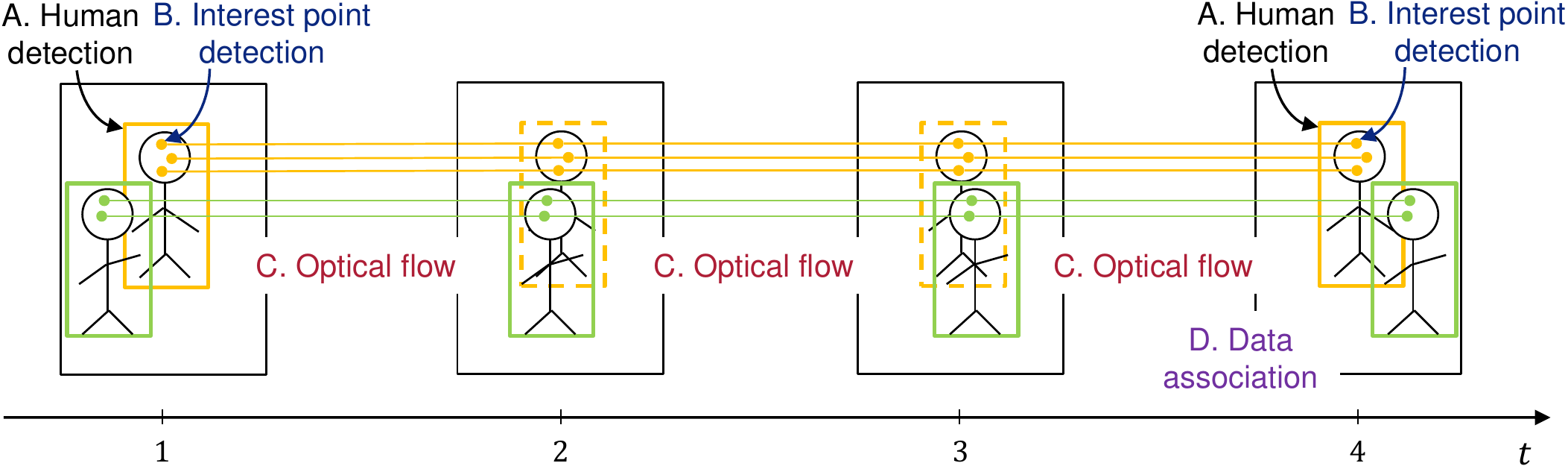}
 \end{center}
 \caption{Human tracking by proposed SDOF-Tracker.}
 \label{fig:proposed}
\end{figure}

First, we formulate the problem of human tracking.
Second, we introduce the overall design of the proposed method, and next, we introduce each module.

\subsection{Problem Formulation}
We formulate the problem of human tracking.
Let $B_t=(\mathbf{b}^1_t, \mathbf{b}^2_t, \cdots)$ be the bounding boxes in frame $\mathbf{o}_t$ at time $t$.
Here, $\mathbf{b}^i_t$ denotes the $i$-th bounding box in frame $\mathbf{o}_t$.
The bounding box is represented in the image coordinate system by $\mathbf{b}=(x, y, w, h)$, where $x$ and $y$ are the top-left $x$ and $y$ coordinates of the bounding box, respectively, and $w$ and $h$ are the width and height of the bounding box, respectively.
For the $i$-th bounding box $\mathbf{b}^i_t$ in frame $\mathbf{o}_t$, let $\mathbf{a}^i_t=(\mathbf{b}^i_t, z^i_t)$ be the human ID $z^i_t$, and let $A_t=(\mathbf{a}^1_t, \mathbf{a}^2_t, \cdots)$ be the collection of all of these in frame $\mathbf{o}_t$.
Human tracking can be formulated as the problem of finding $\{A_t \ |\ t\geq1\}$ given a time series image $\{\mathbf{o}_t \ |\ t\geq1\}$.

\subsection{Overall Design}
We aim to improve the running speed of tracking by using optical flow, which can estimate flow vectors at high speed.
While the high speed tracking is performed using optical flow in every frame, detections are performed at a certain frame interval.
To improve the robustness, interest points are set inside segmented regions and around the head, which is less likely to be occluded.
Moreover, tracking by optical flow is continued for several frames even if a target human is not detected.
This continuation can prevent false negatives, thus also preventing ID switches.

SDOF-Tracker has four modules: A.~human detection, B.~interest point detection, C.~human tracking by optical flow, and D.~data association.
Figure~\ref{fig:proposed} shows human tracking by SDOF-Tracker.
It works in an online manner in that the tracking result is immediately available with each incoming frame.
In the first frame, A.~human detection and B.~interest point detection are performed.
After that, C.~human tracking by optical flow is performed in every frame.
Then, for each $L$ frame (frame $4$ in the figure), A.~human detection, D.~data association, and B.~interest point detection (initialization) are performed.
The details of each module are described from the next section.

\subsection{A. Human Detection} \label{subsec:detection}
This module estimates bounding boxes $D_t=(\mathbf{d}^1_t, \mathbf{d}^2_t, \cdots)$ using the trained human detector.
In this work, we use Mask R-CNN~\cite{he2017mask} not only for detecting humans but also for performing instance segmentation to set robust interest points.
In the first frame, human ID $z^i_t$ is determined to be unique for each $i$.

\subsection{B. Interest Point Detection} \label{subsec:interest}
This module sets interest points inside bounding boxes for optical flow calculation.
In the first frame, target bounding boxes are $D_t=(\mathbf{d}^1_t, \mathbf{d}^2_t, \cdots)$.
On the other hand, in the frame $t (t\geq2,\ t\mid L)$, target bounding boxes are $B_t=(\mathbf{b}^1_t, \mathbf{b}^2_t, \cdots)$.
To improve the robustness, the interest points are set around the head because the head is less likely to be occluded.
We assume that the y-axis of the head is located from $y$ to $y+0.3h$, where $y$ denotes the top-left $y$ coordinates of the bounding box.
Furthermore, we use the instance segmentation result to limit the region of interest points to improve the robustness.
The segmentation region is eroded using the morphological operator~\cite{jain1989fundamentals} to avoid setting interest points in the background regions.
Interest points $P^i_t=(\mathbf{p}^{i1}_t, \mathbf{p}^{i2}_t, \cdots, \mathbf{p}^{iQ}_t)$ are randomly sampled inside the eroded segmentation region, where $Q$ is a predetermined parameter.
We do not use interest point detection methods to reduce the running time.

\subsection{C. Human Tracking by Optical Flow}
In this module, bounding boxes $\widehat{B}_{t}=(\widehat{\mathbf{b}}^1_t, \widehat{\mathbf{b}}^2_t, \cdots)$ are estimated from $B_{t-1}=(\mathbf{b}^1_{t-1}, \mathbf{b}^2_{t-1}, \cdots)$ by optical flow.
In the following, we explain how to predict $i$-th bounding box $\widehat{\mathbf{b}}^i_t$ from the $\mathbf{b}^i_{t-1}$.
First, the optical flow $\Delta^i_t=(\mathbf{\delta}^{i1}_t, \mathbf{\delta}^{i2}_t, \cdots, \mathbf{\delta}^{iQ}_t)$, which indicates where the interest point set $P^i_{t-1}=(\mathbf{p}^{i1}_{t-1}, \mathbf{p}^{i2}_{t-1}, \cdots, \mathbf{p}^{iQ}_{t-1})$ has moved, is estimated.
Second, the location of the bounding box $(\widehat{x}^i_t, \widehat{y}^i_t)$ is obtained by adding the median of the optical flow.
\begin{equation}
(\widehat{x}^i_t, \widehat{y}^i_t) = (x^i_{t-1}, y^i_{t-1}) + \widetilde{\Delta}^i_t
\end{equation}
Third, $w^i_t$ and $h^i_t$ are determined as the same value as time $t-1$ because they change very little between adjacent frames.
Finally, human ID $z^i_t$ inherits the same ID as time $t-1$.

However, the tracking may fail when the interest points track other humans or objects.
In such cases, interest points often spread out rapidly.
In this work, the determination of the termination of tracking is performed using the ratio of the variance of interest points between adjacent frames.
The ratio is calculated by the variance of the interest point $P_{t-1}^i = (\mathbf{p}_{t-1}^{i1}, \mathbf{p}_{t-1}^{i2}, \cdots, \mathbf{p}_{t-1}^{iQ})$ in frame $o_{t-1}$ and the interest point $P_t^i = (\mathbf{p }_{t-1}^{i1}+\mathbf{\delta}_t^{i1}, \mathbf{p}_{t-1}^{i2}+\mathbf{\delta}_t^{i2}, \cdots, \mathbf{p}_{t-1}^{iQ}+\mathbf{\delta}_t^{iQ})$ in frame $o_t$ as follows:
\begin{equation}
\alpha^i_t = \frac{{\rm var}(P_t^i)} {{\rm var}(P_{t-1}^i)}
\end{equation}
Note that the interest points estimated by optical flow may have noise, so we remove such interest points before calculating the variances (\eg Hotelling theory).
Also, the tracking is terminated when the number of interest points becomes less than a predetermined threshold $R$.

\subsection{D. Data Association}
In this module, each bounding box $\{\widehat{\mathbf{b}}^i_t \in \widehat{B}_{t}\}$ estimated by optical flow is associated with each detection $\{\mathbf{d}^i_t \in D_{t}\}$ estimated by the human detector in each $L$ frame.
The data association has three important roles: the estimation of human ID, determination of the start of tracking, and determination of termination of tracking.
The Hungarian algorithm~\cite{kuhn1955hungarian} is used for the association.
The cost matrix for the Hungarian algorithm is calculated using the IoU (Intersection over Union) between the detections and bounding boxes.
When performing association, if the cost is larger than a predefined threshold $\varepsilon$, the bounding box is not associated with the detection to prevent false association.
For each matching pair, human ID $z^i_t$ and bounding box $\mathbf{b}^i_t$ are determined to have the same value as the matched detection $\mathbf{d}^i_t$.
For each unmatched detection, tracking starts as a new human ID.
For each unmatched bounding box, the tracking is terminated. 
However, in crowded scenes, bounding boxes tend to be unmatched due to false negatives.
In this work, even if a bounding box is unmatched within $M$ frames, the tracking is continued.

\section{Experiments}
In order to verify the effectiveness and efficiency of the proposed SDOF-Tracker, human tracking experiments were conducted.

\subsection{Experimental Conditions}
\noindent
{\bf Dataset: }
For the experiments, we used the latest challenging MOT20 dataset~\cite{dendorfer2020mot20}.
MOT20 was captured with a fixed or moving camera in a square, street, and shopping mall.
The frame rate is $25 \mathrm{fps}$, the resolution from ($1173\times880$)--($1,920\times1,080$), the time from $17$--$133$ seconds, and the total number of objects from $90$--$1121$.
We used $4$ sequences in the validation dataset.

\noindent
{\bf Evaluation Metric: }
The evaluation metrics include the number of objects tracked more than $80\%$ of the flow line (Mostly Tracked; MT), the number of objects tracked less than $20\%$ (Mostly Lost; ML), Recall (Rcll), Precision (Prcn), ID switches (IDsw), Fragmentation (Frag), and Multiple Object Tracking Accuracy (MOTA).
MOTA is a widely used and comprehensive metric that combines three error sources (false negative, ID switch, and false positive).
We also measured the average speed per $1$ frame.
We used an Intel Core i7-7700K 4.20GHz CPU, 32GB RAM, and an NVIDIA GeForce Titan X Pascal GPU.

\noindent
{\bf Implementation Details: }
As the baseline method, human detection and feature extraction are performed in every frame.
We used Mask R-CNN~\cite{he2017mask} for the human detector, and it was trained using MS COCO~\cite{lin2014microsoft}.
The same human detection result was used for the baseline and the proposed SDOF-Tracker.
The threshold of human detection was set to $0.2$
%
The following are the parameters for SDOF-Tracker.
The frame interval for human detection was set to $L=5$.
The frame length for tracking continuation was set to $M=10$.
For the optical flow calculation, the Lucas-Kanade method ~\cite{lucas1981iterative} was used.
The max and minimum number of interest points were set to $Q=10$ and $R=3$, respectively.
The parameters for human association were set to $\varepsilon=0.7$.

\subsection{Ablation Study}
\begin{table}[t]
\begin{center}
\scalebox{0.90}{
\begin{tabular}{lll|rrrrrrr}
\toprule
S & C & T & \multicolumn{1}{c}{MT $\uparrow$} & \multicolumn{1}{c}{ML $\downarrow$} & \multicolumn{1}{c}{Rcll [$\%$] $\uparrow$} & \multicolumn{1}{c}{Prcn [$\%$] $\uparrow$} & \multicolumn{1}{c}{IDsw $\downarrow$} & \multicolumn{1}{c}{Frag $\downarrow$} & \multicolumn{1}{c}{MOTA $\uparrow$} \\
\midrule
 &  &  & $228$ & $727$ & $40.7$ & $86.4$ & $13,893$ & $18,319$ & $33.3$ \\
\checkmark &  &  & $226$ & $726$ & $40.7$ & ${\bf 86.6}$ & $13,738$ & $17,932$ & $33.4$ \\
 & \checkmark &  & ${\bf 291}$ & $632$ & $44.8$ & $83.6$ & $9,837$ & $15,422$ & $35.3$ \\
 &  & \checkmark & $220$ & $731$ & $40.5$ & $86.5$ & $14,650$ & $18,968$ & $33.1$ \\
\checkmark & \checkmark & \checkmark & ${\bf 291}$ & ${\bf 623}$ & ${\bf 44.9}$ & $84.1$ & ${\bf 9,578}$ & ${\bf 14,856}$ & ${\bf 35.7}$ \\
\bottomrule
\end{tabular}
}
\end{center}
\caption{Ablation study.
S: Segmentation, C: Continuation, T: Termination.}
\label{tb:ablation}
\end{table}

In this section, we verify the effectiveness of each of the three factors in the SDOF-Tracker, the segmentation for point extraction (S), tracking continuation (C), and tracking termination using interest points (T).
We set the frame interval for human detection to $L=5$.

Table~\ref{tb:ablation} shows the performances with the three factors combined.
First, let us explain the segmentation for point extraction.
As expected, the precision and the number of ID switches improved, and as a result, MOTA improved.
Second, let us explain the tracking continuation.
As expected, recall significantly improved.
As a result, MT, ML, the number of ID switches, the number of fragmentations, and MOTA also improved.
Finally, let us explain the tracking termination using interest points.
Although the precision improved, the number of ID switches and MOTA degraded.
This implies that interest points are not accurately set inside the human region by segmentation.
It is considered that the ratio of the variance of interest points is appropriately calculated because their points are not accurately set inside the human region.
The combination of all of the above achieved the highest performance for almost all metrics (MT, ML, Recall, IDsw, Frag, and MOTA).
In the combination, tracking termination also contributed.
This implies that interest points are accurately set inside the human region by segmentation.

\subsection{Analysis of Accuracy and Speed}
\begin{figure}[t]
 \begin{center}
 \subfloat[MOTA.]{
 \includegraphics[width=0.45\linewidth]{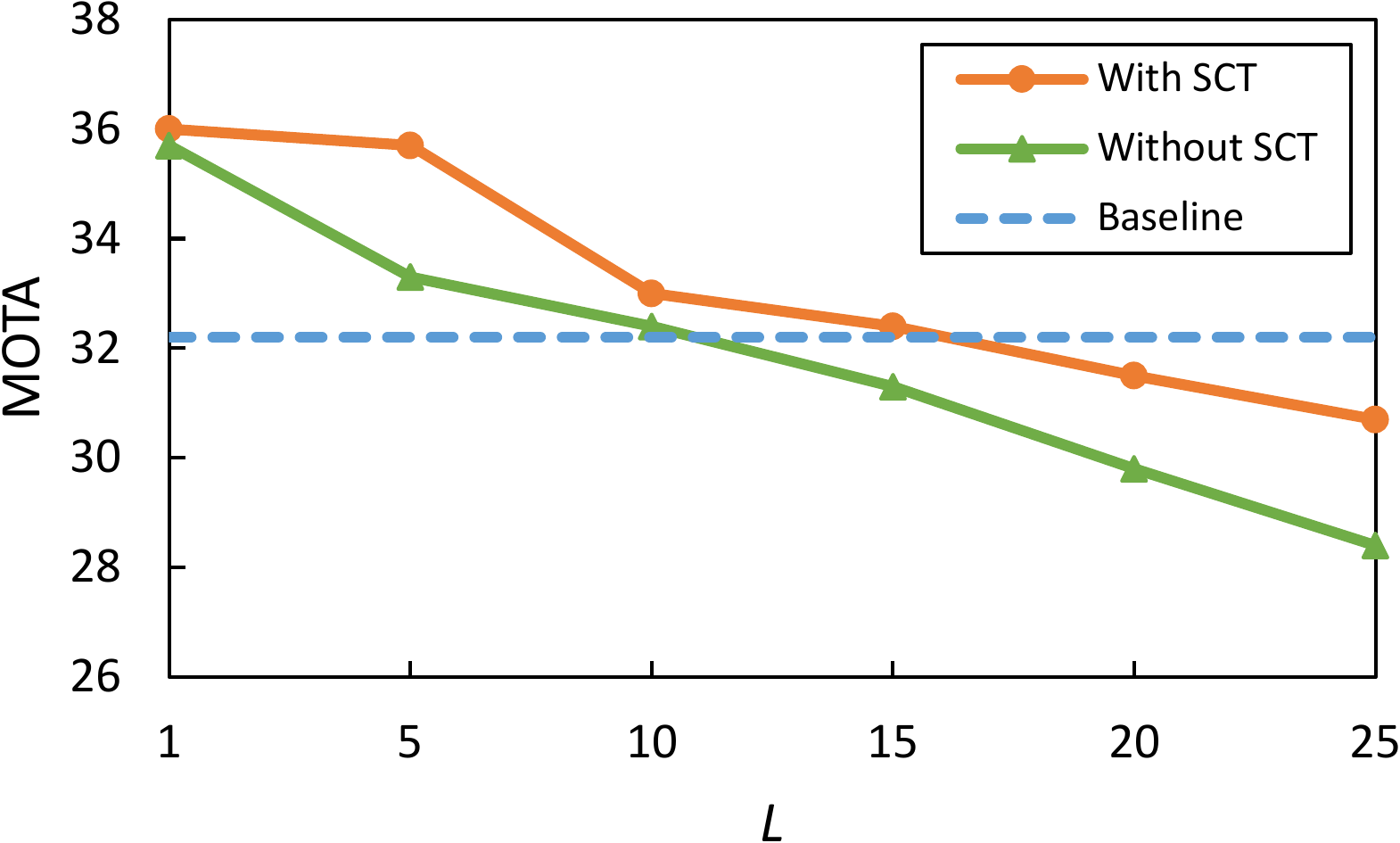}
 \label{fig:mota}
 }
 \subfloat[Speed.]{
 \includegraphics[width=0.45\linewidth]{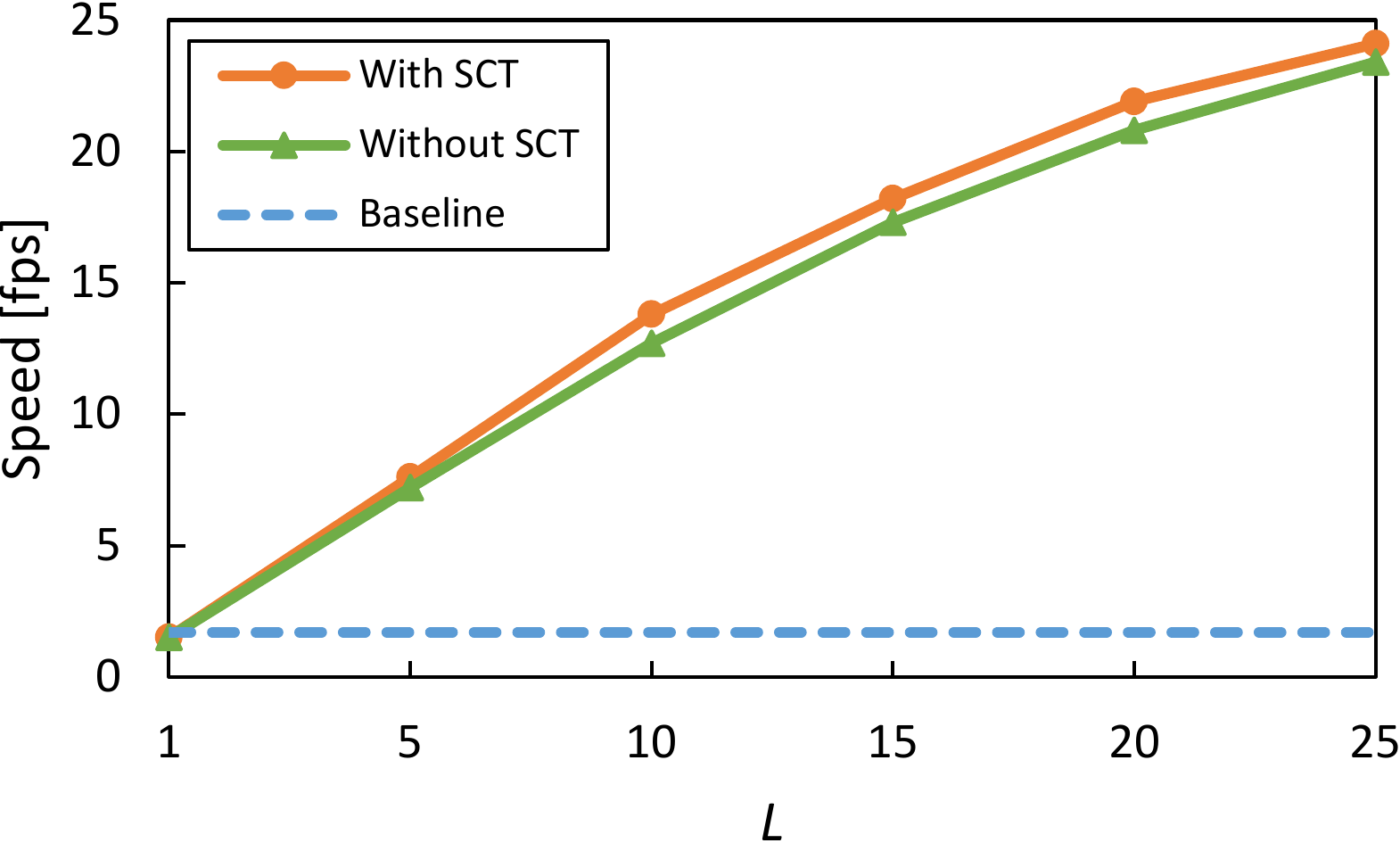}
 \label{fig:speed}
 }
 \end{center}
 \caption{Change in tracking accuracy and speed with increasing frame interval ($L$) for human detection.}
 \label{fig:mota_speed}
\end{figure}

\begin{figure}[t]
 \begin{center}
 \includegraphics[width=0.70\linewidth]{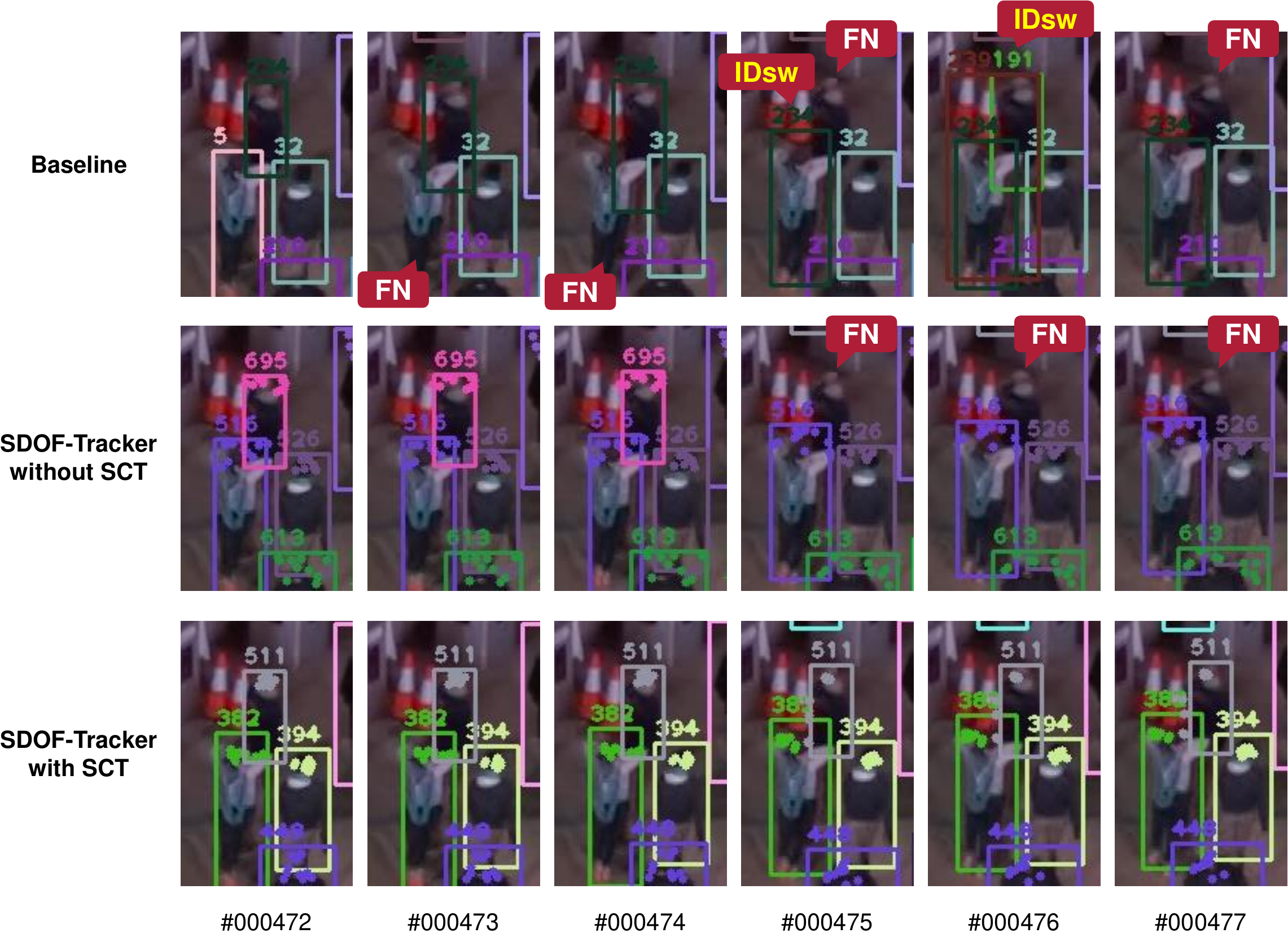}
 \end{center}
 \caption{Cropped example of the tracking result using the baseline and SDOF-Tracker.}
 \label{fig:example}
\end{figure}

\begin{figure}[t]
 \begin{center}
 \includegraphics[width=0.65\linewidth]{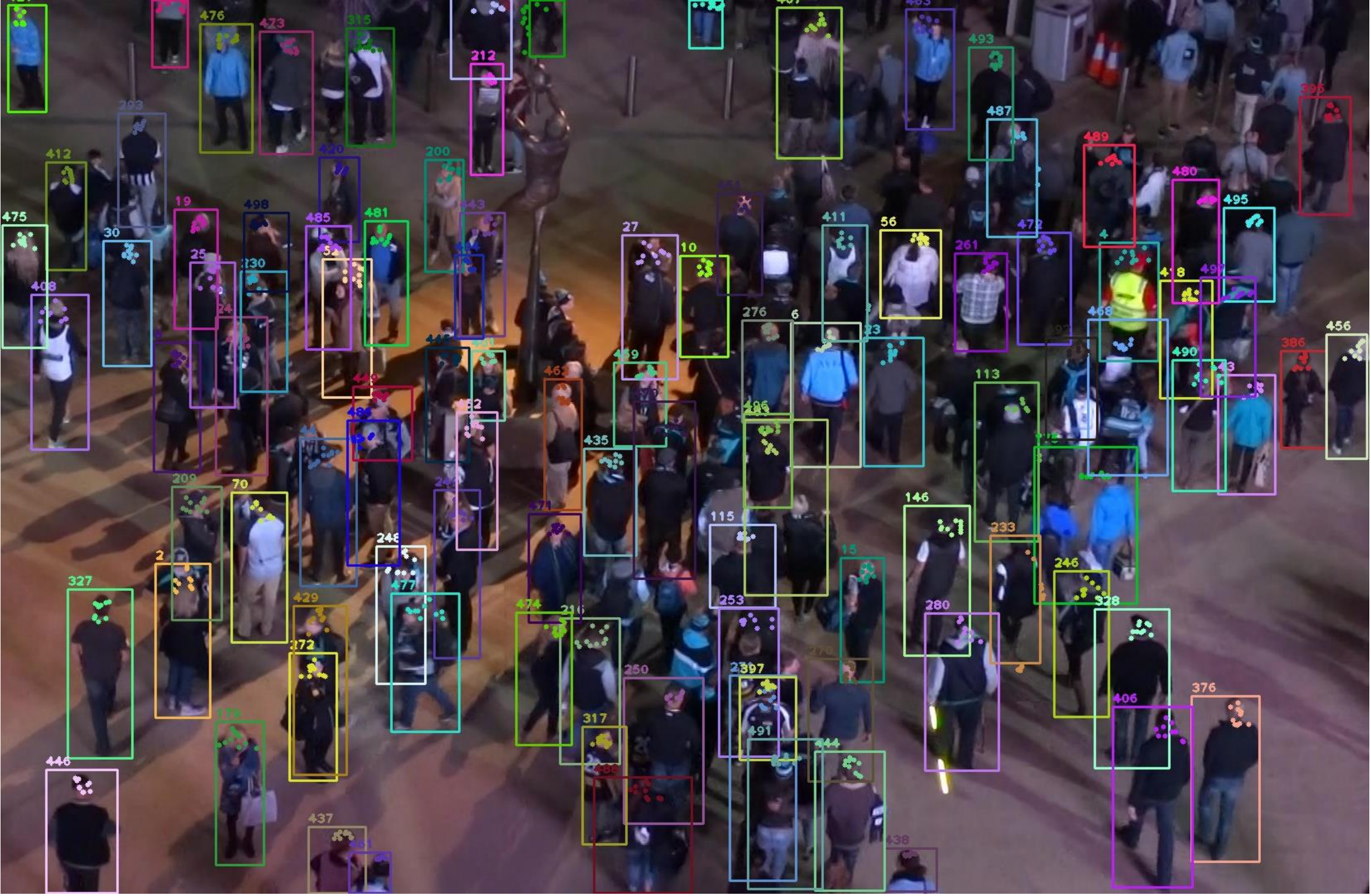}
 \end{center}
 \caption{Example of tracking result using SDOF-Tracker}
 \label{fig:example_overall}
\end{figure}

We evaluated whether the running speed can be improved while maintaining the tracking accuracy when the frame interval ($L$) for human detection is increased.
Speed includes the time required for human detection.
For the baseline method, human detection is performed in every frame, and it is equivalent to DeepSORT~\cite{Wojke2017simple}.
On the other hand, the SDOF-Tracker performs human detection in every $L$ frame.
In SDOF-Tracker, we evaluated whether segmentation, tracking continuation, and tracking termination are performed or not.
In order to compare the accuracy fairly, we use the same detection result using Mask-RCNN in both with/without SCT.
Therefore, the segmentation time is included when evaluating ``without SCT'', but the actual speed without segmentation is even faster.

Figure~\ref{fig:mota} shows the change in the tracking accuracy.
In ``with SCT'', MOTA is almost the same when $L=1$ as when $L=5$.
Then, MOTA decreases when $L\geq5$, and is almost the same when $L=15$ as the baseline.
By contrast, in ``without SCT'', MOTA decreases when $L\geq1$, and is almost the same when $L=10$ as the baseline.
On the other hand, Figure~\ref{fig:speed} shows the change in running speed.
As $L$ increases, the running speed increases in both ``with/without STC''.
The speed improvement rate according to $L$ is higher with STC than without STC.
This is because the frequency of the termination is increased and the number of tracked humans is decreased as $L$ increases.
Thus, SDOF-Tracker with SCT can improve the running speed while maintaining the tracking accuracy.

\subsection{Tracking Examples}
Figure~\ref{fig:example} shows a cropped example of the tracking result using the baseline and SDOF-Tracker.
This is a scene where three people are walking toward the back.
In the baseline, ID switches (IDsw) occur due to false negatives (FN).
In SDOF-Tracker, human detection is performed in frame $475$ because we set $L=5$.
In SDOF-Tracker without SCT, the false negatives are prevented in frame $473$ and $474$ due to tracking by optical flow.
However, the other false negatives remain.
This is because the optical flow cannot start tracking when the false negative occurs in frame $475$, which is a chance for human detection.
On the other hand, in SDOF-Tracker with SCT, all false negatives and ID switches are prevented due to tracking continuation in frame $475$.
Moreover, interest points are accurately set on the regions of human heads.
%
Figure~\ref{fig:example_overall} shows an example of the tracking result using SDOF-Tracker.
Even though this is a very crowded scene, most humans are accurately tracked because interest points are set on the heads.

\subsection{MOTChallenge Result}
\begin{table}[t]
\begin{center}
\scalebox{0.90}{
\begin{tabular}{l|rrrrr|rr}
\toprule
 & \multicolumn{1}{c}{Rcll [$\%$] $\uparrow$} & \multicolumn{1}{c}{Prcn [$\%$] $\uparrow$} & \multicolumn{1}{c}{IDsw $\downarrow$} & \multicolumn{1}{c}{MOTA $\uparrow$} & \multicolumn{1}{c}{Speed [$\mathrm{fps}$] $\uparrow$} & \multicolumn{1}{|c}{Speed [$\mathrm{fps}$] $\uparrow$} & \multicolumn{1}{c}{Speed [$\mathrm{fps}$] $\uparrow$} \\
&&&&& (Tracking) & (Detection) & (Total) \\
\midrule
SDOF-Tracker & $58.0$ & $84.6$ & $3,532$ & $46.7$ & $19.2$ & $38.0$ & $12.8$ \\
SORT~\cite{Bewley2016_sort} & $48.8$ & $90.2$ & $4,470$ & $42.7$ & $57.3$ & $7.6$ & $6.7$ \\
LTSiam~\cite{urbann2021online} & $58.5$ & $84.0$ & $4,509$ & $46.5$ & $30.3$ & $7.6$ & $6.1$ \\
MPNTrack~\cite{braso2020learning} & $61.1$ & $94.9$ & $1,210$ & $57.6$ & $6.5$ & $7.6$ & $3.5$ \\
TBC~\cite{ren2020tracking} & $62.3$ & $89.5$ & $2,449$ & $54.5$ & $5.6$ & $7.6$ & $3.2$ \\
SimpleReID~\cite{karthik2020simple} & $55.3$ & $97.8$ & $2,178$ & $53.6$ & $1.3$ & $7.6$ & $1.1$ \\
Tracktor~\cite{bergmann2019tracking} & $54.3$ & $97.6$ & $1,648$ & $52.6$ & $1.2$ & $7.6$ & $1.0$ \\
TransCenter~\cite{xu2021transcenter} & $71.4$ & $88.3$ & $4,493$ & $61.0$ & $1.0$ & $7.6$ & $0.9$ \\
LPC\_MOT~\cite{dai2021learning} & $58.8$ & $96.3$ & $1,562$ & $56.3$ & $0.7$ & $7.6$ & $0.6$ \\
mfi\_tst~\cite{yang2021online} & $66.6$ & $90.5$ & $1,919$ & $59.3$ & $0.5$ & $7.6$ & $0.5$ \\
GNNMatch~\cite{papakis2021gcnnmatch} & $56.8$ & $96.9$ & $2,038$ & $54.5$ & $0.1$ & $7.6$ & $0.1$ \\
\bottomrule
\end{tabular}
}
\end{center}
\caption{MOTChallenge result on MOT20 dataset.
The result is cited from MOTChallenge web page\protect \footnotemark[1] (Our entry name on the web page is ``FlowTracker'').
}
\label{tb:mot}
\end{table}

We compared the SDOF-Tracker to the state-of-the-art methods in MOTChallenge\footnote[1]{https://motchallenge.net} on the MOT20 dataset.
We compared the performance with methods which have been published in the literature.
We use the public detection results of MOTChallenge to compare both accuracy and speed fairly.
The speed of detection ($7.6\ \mathrm{fps}$) was cited from the literature~\cite{true2021motion}.
Using this speed, the speed of SDOF-Tracker was estimated as $38.0\ \mathrm{fps}$ because the frame interval for human detection was set to $L=5$.
Note that SDOF-Tracker did not use a GPU for human tracking.
Since the public detection results do not include segmentation results, we do not limit regions for setting interest points.
%
Table~\ref{tb:mot} shows the MOTChallenge result.
SDOF-Tracker achieved the best performance in terms of the total speed.
Nevertheless, MOTA was better than SORT and LTSiam.

\section{Conclusion}
In this paper, we proposed SDOF-Tracker, a fast and accurate human tracking method using skipped-detection and optical-flow.
In SDOF-Tracker, tracking by optical flow starts triggered by human detection and ends based on the variance of interest points.
To maintain accuracy, we introduced robust interest point selection within human regions and a tracking termination metric calculated by the distribution of the interest points.
In the experiments, we confirmed that SDOF-Tracker can improve the running speed while maintaining the tracking accuracy when the frame interval for human detection is increased.
Moreover, SDOF-Tracker achieved the best performance in terms of the total speed ($12.8\ \mathrm{fps}$) while maintaining MOTA ($46.7$) on the MOT20 dataset in the MOTChallenge.
In the future, we will develop a method that can dynamically change the frame interval of human detection.

\bibliographystyle{unsrt}  
\bibliography{references}

\end{document}